\documentclass[journal,transmag]{IEEEtran}

\ifCLASSINFOpdf
\else
\fi
\usepackage{times}
\usepackage{epsfig}
\usepackage{graphicx}
\usepackage{amsmath}
\usepackage{amssymb}

\usepackage[]{algorithm2e}

\usepackage[utf8]{inputenc}
\usepackage[english]{babel}
\usepackage{comment}
\usepackage{amsthm}
\usepackage[usenames, dvipsnames]{color}

\usepackage[switch]{lineno}
 
\definecolor{Pink}{rgb}{0.858, 0.188, 0.478}

\begin{document}

%
\title{Auto-context Convolutional Neural Network (Auto-Net) for Brain Extraction in Magnetic Resonance Imaging}


\author{\IEEEauthorblockN{Seyed Sadegh Mohseni Salehi\IEEEauthorrefmark{1,2},~\IEEEmembership{Student Member,~IEEE,}
Deniz Erdogmus\IEEEauthorrefmark{1},~\IEEEmembership{Senior Member,~IEEE}, \\ and
Ali Gholipour\IEEEauthorrefmark{2},~\IEEEmembership{Senior Member,~IEEE}}
\IEEEauthorblockA{\IEEEauthorrefmark{1}Electrical and Computer Engineering Department, Northeastern University, Boston, MA, 02115}
\IEEEauthorblockA{\IEEEauthorrefmark{2}Radiology Department, Boston Children's Hospital; and Harvard Medical School, Boston MA 02115}
\thanks{Manuscript received March 6, 2017.
Corresponding author: S.S.M.Salehi (email: ssalehi@ece.neu.edu). Relevant code can be found at: https://github.com/SadeghMSalehi/AutoContextCNN}}

%



\maketitle
\begin{abstract}
Brain extraction or whole brain segmentation is an important first step in many of the neuroimage analysis pipelines. The accuracy and robustness of brain extraction, therefore, is crucial for the accuracy of the entire brain analysis process. State-of-the-art brain extraction techniques rely heavily on the accuracy of alignment or registration between brain atlases and query brain anatomy, and/or make assumptions about the image geometry; therefore have limited success when these assumptions do not hold or image registration fails. With the aim of designing an accurate, learning-based, geometry-independent and registration-free brain extraction tool in this study, we present a technique based on an auto-context convolutional neural network (CNN), in which intrinsic local and global image features are learned through 2D patches of different window sizes. We consider two different architectures: 1) a voxelwise approach based on three parallel 2D convolutional pathways for three different directions (axial, coronal, and sagittal) that implicitly learn 3D image information without the need for computationally expensive 3D convolutions, and 2) a fully convolutional network based on the U-net architecture. Posterior probability maps generated by the networks are used iteratively as context information along with the original image patches to learn the local shape and connectedness of the brain to extract it from non-brain tissue.

The brain extraction results we have obtained from our CNNs are superior to the recently reported results in the literature on two publicly available benchmark datasets, namely LPBA40 and OASIS, in which we obtained Dice overlap coefficients of 97.73\% and 97.62\%, respectively. Significant improvement was achieved via our auto-context algorithm. Furthermore, we evaluated the performance of our algorithm in the challenging problem of extracting arbitrarily-oriented fetal brains in reconstructed fetal brain magnetic resonance imaging (MRI) datasets. In this application our voxelwise auto-context CNN performed much better than the other methods (Dice coefficient: 95.97\%), where the other methods performed poorly due to the non-standard orientation and geometry of the fetal brain in MRI. Through training, our method can provide accurate brain extraction in challenging applications. This in-turn may reduce the problems associated with image registration in segmentation tasks.

\end{abstract}



\IEEEdisplaynontitleabstractindextext

%
\IEEEpeerreviewmaketitle

\section{Introduction}
%
%
%
%
\IEEEPARstart{W}{hole} brain segmentation, or brain extraction, is one of the first fundamental steps in the analysis of magnetic resonance images (MRI) in advanced neuroimaging applications such as brain tissue segmentation and volumetric analysis \cite{makropoulos2014automatic}, longitudinal and group analysis \cite{li2014mapping}, cortical and sub-cortical surface analysis and thickness measurement~\cite{macdonald2000automated,clouchoux2012quantitative}, and surgical planning. Manual brain extraction is time consuming especially in large-scale studies. Automated brain extraction is necessary but its performance and accuracy are critical as the output of this step can directly affect the performance of all following steps.

Recently neural networks and deep learning have attracted enormous attention in medical image processing. Brebisson et.al.~\cite{de2015deep} proposed the SegNet, a convolutional neural network system to segment different parts of the brain. Recently, CNN-based methods have also been used successfully in tumor segmentation~\cite{wachinger2017deepnat,pereira2016brain,havaei2016brain}, brain lesion segmentation~\cite{brosch2016deep,kamnitsas2016efficient}, and infant brain image segmentation~\cite{zhang2015deep}. In what follows we review the state-of-the-art in whole brain segmentation and the related work that motivated this study. We then introduce a CNN-based method that generates accurate brain extraction.


\section{Related Work}

Many algorithms have been developed and continuously improved over the past decade for whole brain segmentation, which has been a necessary component of large-scale neuroscience and neuroimage analysis studies. As the usage of these algorithms dramatically grew, the demand for higher accuracy and reliability also increased. Consequently, while fully-automated, accurate brain extraction has already been investigated extensively, it is still an active area of research. Of particular interest is a recent deep learning based algorithm~\cite{kleesiek2016deep} that has shown to outperform most of the popular routinely-used brain extraction tools.

The state-of-the-art brain extraction methods and tools use evolved combinations of image registration, atlases, intensity and edge feature information, and level sets/graph cuts to generate brain masks in MRI images. The majority of these algorithms rely heavily on the alignment of the query images to atlases or make strong assumptions about the geometry, orientation, and image features. Yet the outcome of most of these tools is often inaccurate and involves non-brain structures or cuts parts of the brain. Therefore most of these tools offer options and multiple parameters to set and try, that ultimately make brain extraction a semi-automatic or supervised task rather than fully automatic.

Among brain extraction methods four algorithms that are distributed with the widely-used neuroimage analysis software packages, have been evolved and are routinely used. These are the Brain Extraction Tool (BET) from FSL~\cite{smith2002fast,jenkinson2005bet2}, 3dSkullStrip from the AFNI toolkit~\cite{cox1996afni}, the Hybrid Watershed Algorithm (HWA) from FreeSurfer~\cite{lin2003hybrid}, and Robust Learning-Based Brain Extraction (ROBEX)~\cite{iglesias2011robust}. BET expands a deformable spherical surface mesh model initialized at the center-of-gravity of the image based on local intensity values and surface smoothness. 3dSkullStrip, which is a modified version of BET, uses points outside of the expanding mesh to guide the borders of the mesh. HWA uses edge detection for watershed segmentation along with an atlas-based deformable surface model. ROBEX fits a triangular mesh, constrained by a shape model, to the probabilistic output of a brain boundary classifier based on random forests. Because the shape model alone cannot perfectly accommodate unseen cases, Robex also uses a small free-form deformation which is optimized via graph cuts.

The current methods are prone to significant errors when certain geometric assumptions do not hold, features are not precisely identified, or image registration, which is often not guaranteed to converge to an exact solution, fails. The problems associated with registration-based segmentation, and the recent promising results in neural network based image segmentation motivate further development and use of learning-based, geometry-independent, and registration-free brain image segmentation.

Recently, Kleesiek et. al.~\cite{kleesiek2016deep} proposed a deep learning based algorithm for brain extraction, which will be referred to as PCNN in this paper. PCNN uses seven 3D convolutional layers for voxelwise image segmentation. Cubes of size $53\times53\times53$ around the grayscale target voxel are used as inputs to the network. In the extensive evaluation and comparison reported in~\cite{kleesiek2016deep}, PCNN outperformed state-of-the-art brain extraction algorithms in publicly available benchmark datasets.

In this study we introduce auto-context CNNs with two network architectures to significantly improve brain extraction accuracy. In our first network, which is a voxelwise architecture, instead of using 3D convolutional layers with one window size (used in PCNN), we use 2D patches of three different sizes as proposed by Moeskops et al.~\cite{moeskops2016automatic}. In addition, to account for 3D structure, and efficiently learn from 3D information to identify brain voxels from non-brain voxels, we use three parallel pathways of 2D convolutional layers in three planes (i.e. axial, coronal and sagittal planes). Our second architecture is a U-net~\cite{ronneberger2015u} style network, in which we use a weighted cost function to balance the number of samples of each class in training. We discuss the details of our proposed auto-context networks, generally referred to as Auto-Net, in this paper.

Context information has shown to be useful in computer vision and image segmentation tasks. Widely-used models, such as conditional random fields~\cite{lafferty2001conditional}, rely on fixed topologies thus offer limited flexibility; but when integrated into deep CNNs, they have shown significant gain in segmentation accuracy~\cite{chen2014semantic,kamnitsas2016efficient}. To increase flexibility and speed of computations, several cascaded CNN architectures have been proposed in medical image segmentation~\cite{wachinger2017deepnat,havaei2016brain,valverde2017improving}. In such networks, the output layer of a first network is concatenated with input to a second network to incorporate spatial correspondence of labels. To learn and incorporate context information in our CNN architectures, we adopt the auto-context algorithm~\cite{tu2010auto}, which fuses low-level appearance features with high-level shape information. As compared to a cascaded network, an auto-context CNN involves a generic and flexible procedure that uses posterior distribution of labels along with image features in an iterative supervised manner until convergence. To this end, the model is flexible and the balance between context information and image features is naturally handled.

Experimental results in this study show that our Auto-Net methods outperformed PCNN and the four widely-used, publicly-available brain extraction techniques reviewed above on two benchmark datasets (i.e. LPBA40 and OASIS, described in Section IV.A). On these datasets we achieved significantly higher Dice coefficients by the proposed Auto-Nets compared to the routinely-used techniques, as auto-context significantly boosted sensitivity while improving or maintaining specificity. We also examined the performance of the Auto-Net in the challenging problem of extracting fetal brain from reconstructed fetal brain MRI. In this case we only compared our results to BET and 3dSkullStrip as the other methods were not designed to work with the non-standard orientation and geometry of the fetal brain in MRI. We present the methods, including the network architectures and the auto-context CNN, in the next section and follow with experimental results in Section IV and a discussion in Section V.

\section{Method}
\subsection{Network Architecture}
We design and evaluate two Auto-Nets with two different network architectures: 1) a voxelwise CNN architecture~\cite{ciresan2012deep}, and 2) a fully convolutional network~\cite{long2015fully, shelhamer2017fully} based on the U-net architecture~\cite{ronneberger2015u}. We describe the details of the network architectures here and follow with our proposed auto-context CNN algorithm in the next subsection.

\subsubsection{A voxelwise network}
\label{voxelbased}

The proposed network has nine types of input features and nine corresponding pathways which are merged in two levels. Each pathway contains three convolutional layers. This architecture segments a 3D image voxel-by-voxel. For all voxels in the 3D image three sets of in-plane patches in axial, coronal, and sagittal planes are used. Each set contains three patches with window sizes of $15\times15$, $25\times25$ and $51\times51$. By using these sets of patches with different window size, both local and global features of each voxel are considered during training. Network parameters are learned simultaneously based on orthogonal-plane inputs, so 3D features are learned without using 3D convolution which is computationally expensive.
\begin{figure*}
    \centering
    \includegraphics[width=\textwidth]{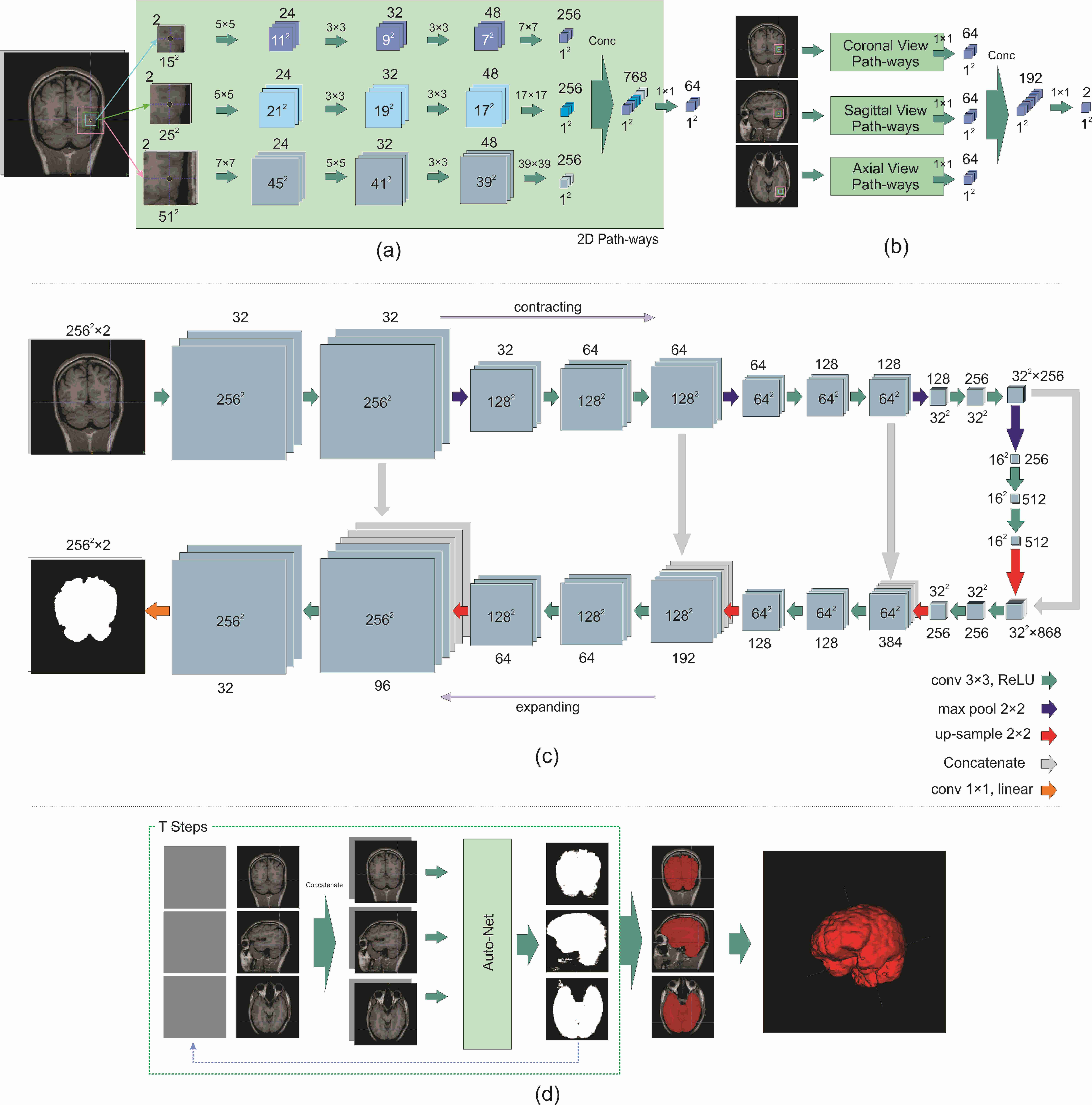}
    \caption{Schematic diagram of the proposed networks: a) The proposed voxelwise architecture for 2D image inputs; b) the network architecture to combine the information of 2D pathways for 3D segmentation; c) the U-net style architecture. The 2D input size for the LPBA40 and fetal MRI datasets was $256\times 256$ and for the OASIS dataset was $176\times 176$; and d) the auto-context formation of the network to reach the final results using network (a) as example. The context information along with multiple local patches are used to learn local shape information from training data and predict labels for the test data.}
    \label{fig:Net}
\end{figure*}

Figure~\ref{fig:Net}(a) shows the schematic architecture of the parallel 2D pathways for one of the 2D views. In the first layer, 24 $5 \times 5$ kernels for the patches of size $15 \times 15$ and $25 \times 25$, and $7 \times 7$ kernels for the patches of size $51 \times 51$ are used. After the first convolutional layer, ReLU nonlinear function and batch normalization is applied. For the second convolutional layer, ReLU nonlinear function is used after applying convolutional layer with 32 convolutional kernels of sizes $3 \times 3$, $3 \times 3$ and $5 \times 5$, for each patch, respectively. In the last convolutional layer 48 kernels of size $3 \times 3$ are used. In the proposed architecture, fully convolutional layers are used instead of fully connected layers~\cite{sermanet2013overfeat} to achieve much faster testing time, as the whole image can be tested in a network with convolutional layers while voxels are tested in a network with fully connected layers. After applying ReLU function, the output of the third convolutional layer is connected to a convolution-type, fully-connected layer with 256 kernels. Then, the nodes for each patch are concatenated and a $1\times1$ convolution with 64 kernels is applied. Each of the 2D pathways collects the information of a 2D plane.

To combine the information of 2D planes, the outputs of each set of in-plane patches are concatenated. This results in 192 nodes in total. Two kernels of $1\times1$ convolutional layers (for brain and non-brain classes) are applied on concatenated nodes with a softmax output layer. Figure~\ref{fig:Net}(b) illustrates this step. We refer to this combination of three 2D pathways network as our 2.5D-CNN. By adding the auto-context algorithm to this architecture (Auto-2.5D-CNN), we aim to combine low-level features from patches with context information learned by the network to improve classification accuracy.

\subsubsection{A fully convolutional network}

The voxelwise approach has two drawbacks: 1) although using fully convolutional layers instead of fully connected layers makes the algorithm faster, it is still relatively slow; and 2) there is a trade-off between finding local features and global features that involves choosing the window size around voxels. In the previous section we described how we conquered the latter problem by choosing different window sizes. Nonetheless, these drawbacks can also be addressed by using a fully convolutional network (FCN)~\cite{long2015fully}. To this end, we use the U-net~\cite{ronneberger2015u} which consists of a contracting path that captures global features and an expanding path that enables precise localization.

The U-net style architecture is shown in Figure~\ref{fig:Net}(c). This architecture consists of a contracting path (to the right) and an expanding path (to the left). The contracting path contains padded $3\times3$ convolutions followed by ReLU non-linear layers. A $2\times2$ max pooling operation with stride 2 is applied after every two convolutional layers. After each downsampling by the max pooling layers, the number of features is doubled. In the expanding path, a $2\times2$ upsampling operation is applied after every two convolutional layers, and the resulting feature map is concatenated to the corresponding feature map from the contracting path. At the final layer a $1\times1$ convolution with linear output is used to reach the feature map with a depth equal to the number of classes (brain or non-brain tissue). We refer to this network as the U-net, as we aim to augment it with the auto-context algorithm (Auto-U-net).

\subsection{Auto-Context CNN}
We propose auto-context convolutional neural networks by adopting the auto-context algorithm developed in~\cite{tu2010auto}. Assuming $m$ training image pairs $\{ (X^{(j)}, Y^{(j)}), j = 1...m\}$, each 3D image is flattened into a 1D vector $X^{(j)} = (x^{(j)}_{1}, x^{(j)}_{2}, ..., x^{(j)}_{n})$ and its corresponding label image is flattened into the vector $Y^{(j)} = (y^{(j)}_{1}, y^{(j)}_{2}, ..., y^{(j)}_{n})$; where $y^{(j)}_{i} $ is the label of voxel $i$ in image $j$. In each image the posterior probability of voxel $i$ having label $l$, computed through a CNN $f_{y_i}(.)$, by the softmax classifier can be written as:
\begin{equation}
    p(y_i=l|X(N_i)) = \frac{e^{f_{y_l}(N_i)}}{\sum_ce^{f_{y_c}(N_i)}}
    \label{softmax}
\end{equation} 
where $N_i$ is the set of patches around voxel $i$, $i=1,...,n$, and $c$ is the number of classes ($l=0,...,c-1$). During the optimization, the cross-entropy between the true distribution $q$ and the estimated distribution $p$, i.e. $H(q,p) = -\sum_i\ q(y_i)\mathrm{log}~p(y_i|X(N_i))$, is minimized. The true distribution follows the Dirac function, i.e. $q(y_i)$ is 1 for the true label and 0 otherwise. The cost function, therefore, would be:
\begin{equation}
    H = -\sum_i \mathrm{log}~p(y_i = true Label|X(N_i)) 
    \label{costFunction}
\end{equation}

In auto-context CNN, a sequence of classifiers is designed in a way that, to train each classifier, the posterior probabilities computed by the previous classifier are used as features. More specifically, for each image at step $t$ the pair of ${X(N_i), p_{(t-1)}(N_i)}$ is considered as a feature for classification of voxel $i$, where $p_{(t-1)}(N_i)$ is the posterior probability of voxels around voxel $i$. Algorithm~\ref{algorithm} shows how the sequence of weights in the network are computed for the sequence of classifiers. The learned weights are used at test time for classification. The proof of convergence of Algorithm~\ref{algorithm} is shown in Appendix~\ref{appendix}. 

\begin{algorithm}

 The training MRI image pairs $\{ (X^{(j)}, Y^{(j)}), j = 1...m\}$ construct uniform distribution of $p^{(j)}_0(N_i)$ on the labels\;
 \Repeat{$I < \epsilon$ }{
  Make a training set $S_{(t)} = \{(y^{(j)}_{i},(X^{(j)}(N_i),p^{(j)}_{(t-1)}(N_i)), j=1...m, i=1...n\}$\; 
  
  Train CNN network using architecture described in figure~\ref{fig:Net}~(a,b: for voxel-wised, c: for FCN)\;
  
  Calculate $p^{(j)}_{(t)}(N_i)$ for $\{j = 1...m, i = 1...n\}$\ using (\ref{softmax})\;
  Calculate $H_t$ using (\ref{costFunction})\; 
  
  $I = |H_{(t)} - H_{(t-1)}|$\;
  
 }
 \caption{The auto-context CNN algorithm}
 \label{algorithm}
\end{algorithm}

To illustrate more on the effect of the auto-context algorithm, consider the first convolutional layer of each 2D pathway in the 2.5D-CNN. Suppose $y$ is an input 3D patch result of concatenating the predicted label and data patches, and $x$ is the output of the first layer for one of the kernels. For the convolution operation with kernel size $k$ we have
\begin{equation}
    x = \sum_{i=1}^dW_{i} * y_i + b
    \label{eq:1}
\end{equation}
where $W$ is a $k\times k\times d$ weight matrix, $*$ is the 2D convolution operation, $d$ is the depth of the input feature which is 2, and $b$ is the bias. Expanding the summation in equation~(\ref{eq:1}) we have
\begin{equation}
    x = W_{1} * y_1 + W_{2} * y_2 + b
    \label{eq:2}
\end{equation}
where $W_1$ and $W_2$ are $k\times k$ weight matrices corresponding to the intensity input ($y_1$) and label input ($y_2$), respectively. $W_2$ values are optimized such that they encode information regarding the shape of the brain labels, their respective location, and the connectedness of the labels. During the training of the network at step 0, the weights corresponding to the label input, $W_2$, are assigned much lower values than the weights corresponding to the intensity input (i.e. $W_2<<W_1$) since the label input carries no information about the image at the beginning. Note that  $p_j^0(N_i)$ is constructed with uniform distribution over classes. On the other hand, in the following steps, the weights corresponding to the label input, $W_2$, are assigned higher values than the weights corresponding to the intensity input (i.e. $W_2>W_1$). Consequently, in testing, the filters corresponding to the predicted labels are more effective than the filters corresponding to intensities. 

\subsection{Training}
\subsubsection{Voxelwise network}
MRI image labels are often unbalanced. For brain extraction the number of non-brain voxels is on average roughly 10 times more than the number of brain voxels. The following process was used to balance the training samples: for each training image, 15000 voxels were randomly selected such that 50\% of the training voxels were among border voxels. The voxels which had two different class labels in a cube of five voxels around them were considered border voxels. Of the remaining 50\% of samples, 25\% were chosen randomly from the brain class and 25\% were chosen from the non-brain class.

For training, the cross-entropy loss function was minimized using ADAM optimizer~\cite{kingma2014adam}. Three different learning rates were employed during the training: In the first step, a learning rate of 0.001 was used with 5000 samples for each MRI data pair and 15 epochs. In the second step, learning rate of 0.0001 was used to update the network parameters with another 5000 samples for each MRI data and 15 epochs. Finally, the last 5000 samples for each MRI data were used with a learning rate of 0.00005 to update the network parameters.  The total training time for this architecture was less than two hours.

\subsubsection{Fully convolutional network}

The output layer in the FCN consists of $c$ planes, one per class ($c=2$ in brain extraction). We applied softmax along each pixel to form the loss. We did this by reshaping the output into a  $ width\times height\times c$ matrix and then applying cross entropy. To balance the training samples between classes we calculated the total cost by computing the weighted mean of each class. The weights are inversely proportional to the probability of each class appearance, i.e. higher appearance probabilities led to lower weights. Cost minimization on 15 epochs was performed using ADAM optimizer~\cite{kingma2014adam} with an initial learning rate of 0.001 multiplied by 0.9 every 2000 steps. The training time for this network was approximately three hours on a workstation with an Nvidia Geforce GTX1080 GPU.

Figure \ref{fig:Net}d illustrates the procedure of using Algorithm~\ref{algorithm}. To create patches for each voxel in the network, two sets of features are used; first, patches of different sizes around each voxel are considered as inputs, i.e. $X(N_i)$. Second, exact same patch windows are considered around the posterior probability maps calculated in the previous step, $p_j^{t-1}(N_i)$, as additional sets of inputs. The posterior probabilities are multiplied to the mean of the data intensity to be comparable with data intensities. Concatenating these two 2D features provides 3D inputs to the network in two different domains.

Training was stopped when it reached convergence, i.e. when the change in the cross-entropy cost function became asymptotically smaller than a predefined threshold $\epsilon$:
\begin{equation}
    I_t=|H_{(t)}-H_{(t-1)}|<\epsilon
    \label{eq:convergence}
\end{equation}
For testing, the auto-context algorithm was used with two steps.\\

\section{Experiments}
\subsection{Datasets}
We evaluated our algorithm first on two publicly available benchmark datasets and then on fetal MRI data which exhibits specific challenges such as non-standard, arbitrary geometry and orientation of the fetal brain, and the variability of structures and features that surround the brain. We used two-fold cross-validation in all experiments. The output of all algorithms was evaluated against the ground truth which was available for the benchmark datasets and was manually obtained prior to this study for the fetal MRIs.

The first dataset came from the LONI Probabilistic Brain Atlas Project (LPBA40) \cite{shattuck2009online}. This dataset consists of 40 T1-weighted MRI scans of healthy subjects with spatial resolution of $0.86 \times 1.5 \times 0.86$ mm. The second dataset involved the first two disks of the Open Access Series of Imaging Studies (OASIS) \cite{marcus2007open}. This consisted of 77 $1 \times 1 \times 1$ mm T1-weighted MRI scans of healthy subjects and subjects with Alzheimer's disease. 

The third dataset contained 75 reconstructed T2-weighted fetal MRI scans. Fetal MRI data was obtained from fetuses scanned at a gestational age between 19 and 39 weeks (mean=30.1, stdev=4.6) on 3-Tesla Siemens Skyra scanners with 18-channel body matrix and spine coils. Repeated multi-planar T2-weighted single shot fast spin echo scans were acquired of the moving fetuses, Ellipsoidal brain masks defining approximate brain regions and bounding boxes in the brain region were defined in ITKSNAP~\cite{yushkevich2006user}, and the scans were then combined through robust super-resolution volume reconstruction by either of the algorithms developed in \cite{gholipour2010robust} or \cite{kainz2015fast} for motion correction and volume reconstruction at isotropic resolution of either 0.75 or 1 mm. Brain masks were manually drawn on the reconstructed images by two experienced segmenters. Manual brain extraction took between 1 to 4 hours per case depending on the age and size of the fetal brain and the quality of the images.

\subsection{Results}
To evaluate the performance of the algorithms, Dice overlap coefficient was used to compare the predicted brain mask $P$ with ground truth mask (extracted manually) $R$. The Dice coefficient was calculated as follow:
\begin{equation}
    D = \frac{2\left | P\cap R \right |}{\left | P \right |+\left | R \right |} = \frac{2TP}{2TP+FP+FN}
\end{equation}
where $TP$, $FP$, and $FN$ are the true positive, false positive, and false negative rates, respectively. We also report specificity, $\frac{TN}{TN+FP}$, and sensitivity, $\frac{TP}{TP+FN}$, to compare algorithms.

\begin{figure}
    \centering
    \includegraphics[width=\columnwidth]{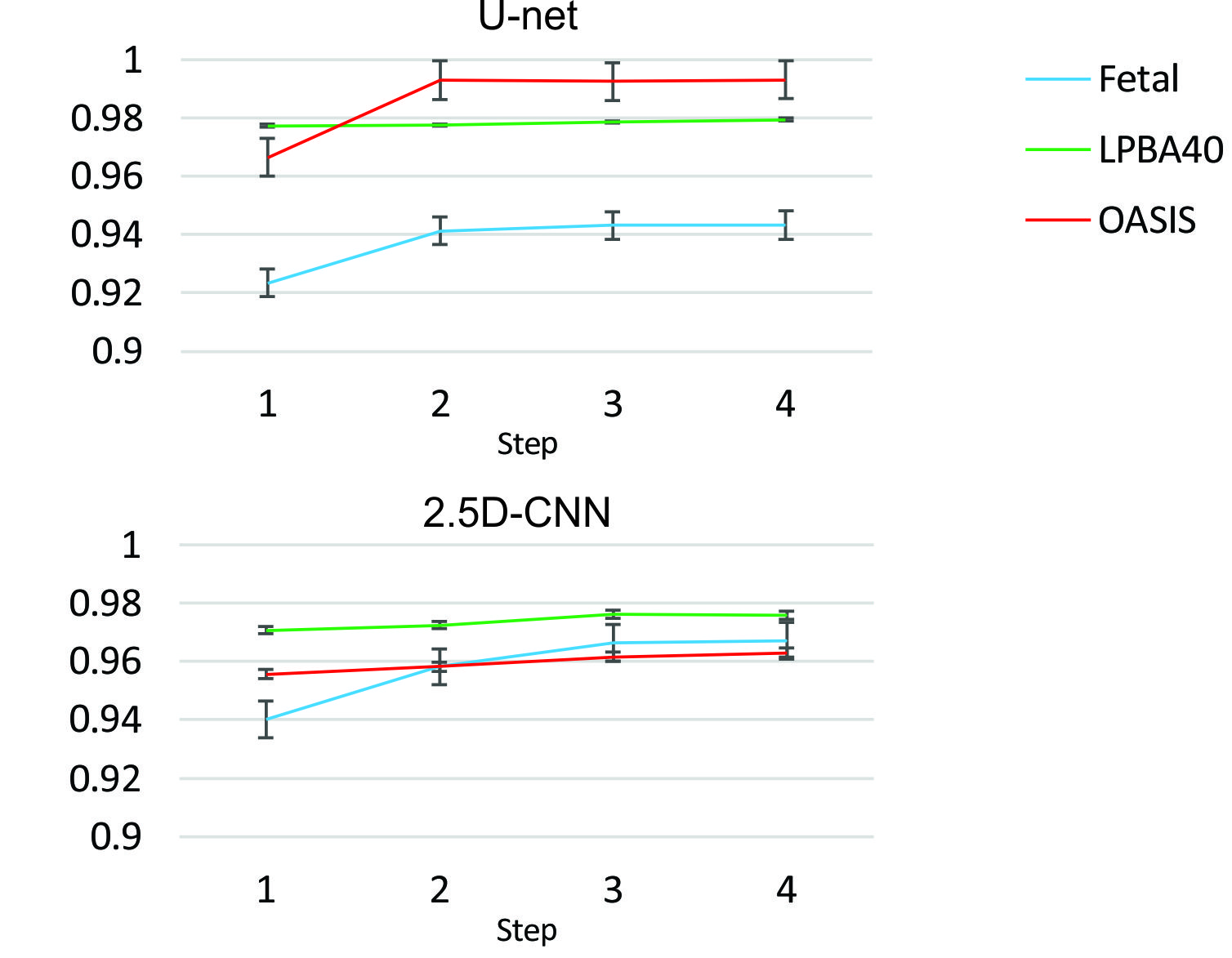}
    \caption{The Dice coefficient of training at four steps of the auto-context algorithm on all datasets based on the U-net (up) and the voxelwise 2.5D CNN approach (bottom). These plots show that the networks learned the context information through iterations and they converged.}
    \label{fig:Train}
\end{figure}

Figure~\ref{fig:Train} shows the Dice coefficient for the different steps of the training session for all datasets in the auto-context CNN algorithm. Improvement in the Dice coefficient is observed in both network architectures (U-net and 2.5D-CNN) through the steps of the auto-context algorithm.

Table~\ref{table:Public} shows the results of our proposed method compared to the other methods on the two benchmark datasets. The results for PCNN were taken from~\cite{kleesiek2016deep}. Auto-context CNNs (Auto-Nets) showed the highest Dice coefficients among all methods, with an increase of about 0.8\% over the best performing methods in the LPBA40 dataset. This significant boost in performance was achieved in Auto-Nets through the auto-context algorithm which, by incorporating local shape context information along with local patches, allowed a significant increase in sensitivity and an increase in specificity.

\begin{table*}[h!]
\small
\centering
 \begin{tabular}{|c||c|c|c||c|c|c|} 
 \hline
  &  \multicolumn{3}{c||}{LPBA40} & \multicolumn{3}{c|}{OASIS}\\
 \hline
Method & Dice & Sensitivity & Specificity & Dice & Sensitivity & Specificity \\
 \hline
Auto-U-net & \textbf{97.73} ($\pm0.003$) & 98.31 ($\pm0.006$) & \textbf{99.48} ($\pm0.001$)& \textbf{97.62} ($\pm0.01$) & \textbf{98.66} ($\pm0.01$) & 98.77 ($\pm0.01$) \\ 
 \hline
U-net & 96.79 ($\pm0.004$) & 97.22 ($\pm0.01$) & 99.34 ($\pm0.002$)& 96.22 ($\pm0.006$) & 97.29 ($\pm0.01$) & 98.27 ($\pm0.007$) \\ 
 \hline
Auto-2.5D-CNN & 97.66 ($\pm0.01$) & 98.25 ($\pm0.01$) & 99.47 ($\pm0.002$)& 96.06 ($\pm0.007$) & 96.21 ($\pm0.01$) & 98.56 ($\pm0.006$) \\ 
 \hline
2.5D-CNN & 97.17 ($\pm0.005$) & 98.52 ($\pm0.01$) & 99.24 ($\pm0.002$)& 95.61 ($\pm0.007$) & 96.3 ($\pm0.01$) & 98.20 ($\pm0.01$) \\ 
 \hline
PCNN & 96.96 ($\pm0.01$) & 97.46 ($\pm0.01$) & 99.41 ($\pm0.003$)& 95.02 ($\pm0.01$) & 92.40 ($\pm0.03$) & \textbf{99.28} ($\pm0.004$) \\ 
\hline
BET & 94.57 ($\pm0.02$) & 98.52 ($\pm0.005$) & 98.22 ($\pm0.01$)& 93.44 ($\pm0.03$) & 93.41 ($\pm0.04$) & 97.70 ($\pm0.02$) \\
\hline
Robex & 95.40 ($\pm0.04$) & 94.25 ($\pm0.05$) & 99.43 ($\pm0.004$)& 95.33 ($\pm0.01$) & 92.97 ($\pm0.02$) & 99.21 ($\pm0.004$) \\
\hline
3dSkullStrip & 92.99 ($\pm0.03$) & 96.95 ($\pm0.01$) & 97.87 ($\pm0.01$)& 92.77 ($\pm0.01$) & 94.44 ($\pm0.04$) & 96.82 ($\pm0.01$) \\ 
\hline
HWA & 92.41 ($\pm0.007$) & \textbf{99.99} ($\pm0.0001$) & 97.07 ($\pm0.004$)& 94.06 ($\pm0.01$) & 98.06 ($\pm0.01$) & 96.34 ($\pm0.01$) \\ 
 \hline
\end{tabular}
\caption{Mean and standard deviation of the scores for different algorithms on LPBA40 and OASIS datasets. The results show that our algorithm increased both sensitivity and specificity and resulted in highest Dice scores among all widely-used tools and the recent PCNN method \cite{kleesiek2016deep}.}
\label{table:Public}
\end{table*}

\begin{figure*}
\centering
\includegraphics[width=\textwidth]{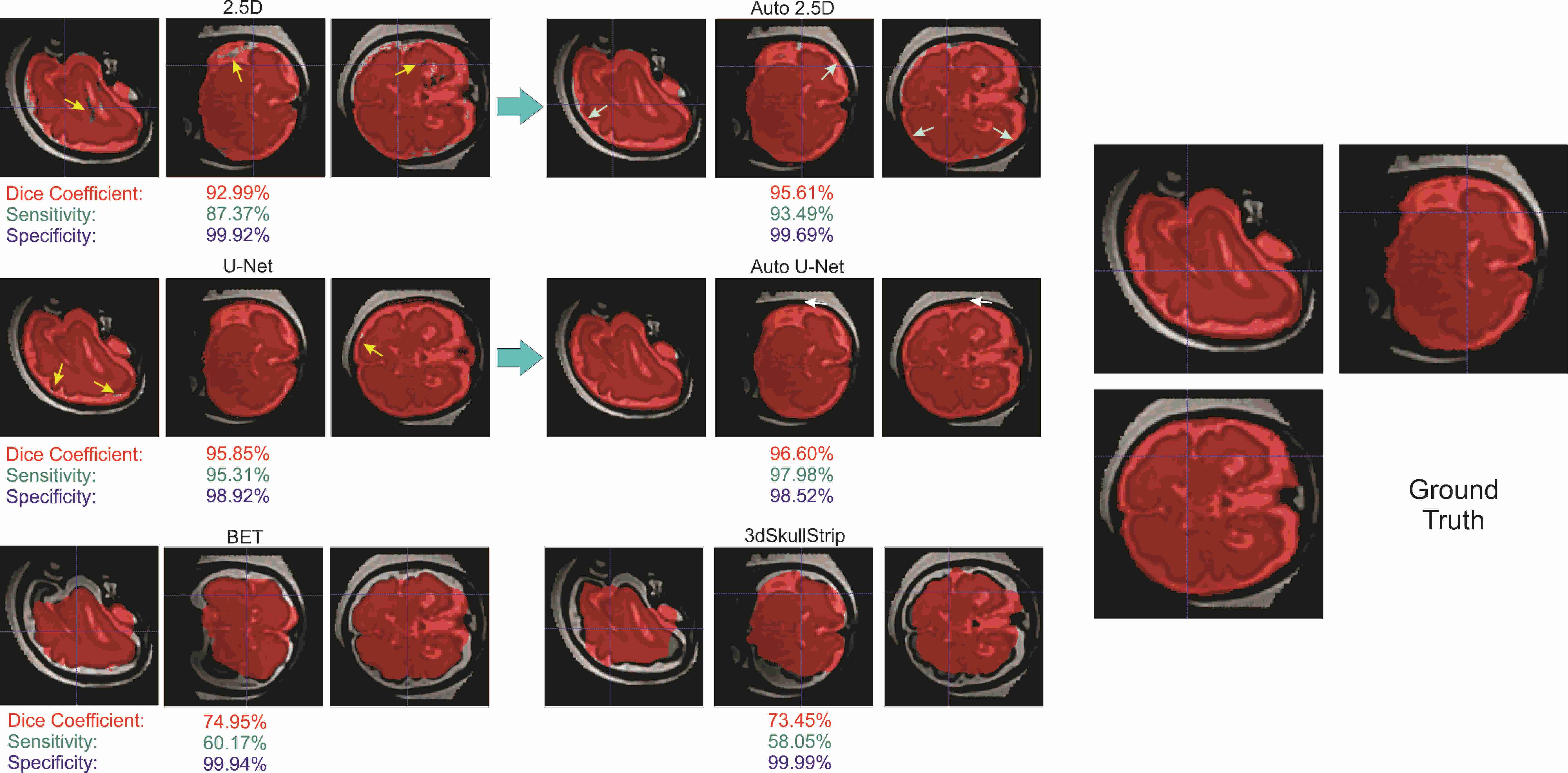}
\caption{Predicted masks overlaid on the data for fetal brain MRI; the top images show the improvement of the predicted brain mask in different steps of the Auto-Net using 2.5D-CNN. The middle images show the improvement of the predicted brain mask in different steps of the Auto-Net using U-Net. The bottom left and right images show the predicted brain masks using BET and 3dSkullStrip, respectively. The right image shows the ground truth manual segmentation. Despite the challenges raised, our method (Auto-Net) performed very well and much better than the other methods in this application. The Dice coefficient, sensitivity, and specificity, calculated based on the ground truth for this case, are shown underneath each image in this figure.}
\label{fig:FetalSample}
\end{figure*}

The main advantage of our CNN-based method was revealed in the fetal MRI application where the fetal brains were in different orientations and surrounded by a variety of non-brain structures. Figure~\ref{fig:FetalSample} shows an example, and Table~\ref{table:Fetal} shows the results of whole brain segmentation on reconstructed fetal MRI. Only Auto-Net styles, BET and 3dSkullStrip were included in this comparison as the other methods were not designed to work with arbitrary brain orientation in fetal MRI and thus performed poorly. As expected, the auto-context algorithm improved the results significantly, and the Auto-Nets performed much better than the other algorithms in this application, with average Dice coefficients that were more than 12\% higher than the other techniques, and sensitivities that were higher by a margin of more than 20\%. In fact, as seen in Figure~\ref{fig:FetalSample}, the other two algorithms generated conservative brain masks which resulted in high specificity (close to 1) but very low sensitivity. The Dice coefficient, sensitivity, and specificity, calculated based on the ground truth for this case, are shown underneath each image in this figure.

\begin{table}[h!]
\small
\centering
 \begin{tabular}{|c||c|c|c|} 
 \hline
Method & Dice & Sensitivity & Specificity \\
 \hline
Auto-U-net & 93.80($\pm0.02$) & 94.64($\pm0.04$) & 98.65($\pm0.01$) \\ 
 \hline
U-net & 92.21($\pm0.03$) & \textbf{96.46}($\pm0.03$) & 97.57($\pm0.01$) \\ 
 \hline
{\footnotesize Auto-2.5D-CNN} &\textbf{95.97}($\pm0.02$) & 94.63($\pm0.02$) & 99.53($\pm0.004$) \\ 
 \hline
2.5D-CNN & 94.01($\pm0.01$) & 94.20($\pm0.03$) & 98.88($\pm0.008$) \\ 
 \hline
BET & 83.68($\pm0.07$) & 73.00($\pm0.1$) & 99.91($\pm0.001$) \\
\hline
3dSkullStrip & 80.57($\pm0.12$) & 69.19($\pm0.16$) & \textbf{99.97}($\pm0.001$) \\
\hline
\end{tabular}
\caption{Mean and standard deviation of the scores of different algorithms on the fetal dataset. The results show that highest Dice coefficients were obtained by Auto-Net compared to BET and 3dSkullStrip among the techniques that could be used in this application. Also, the voxelwise approach (Auto-2.5D-CNN) performed much better than the FCN (Auto-U-net) in this application.}
\label{table:Fetal}
\end{table}

The effect of using the auto-context algorithm can also be seen in Figure~\ref{fig:FetalSample}, where the voxelwise and fully convolutional networks on the right (i.e. 2.5D and U-Net, respectively) are the networks without auto-context. Three different improvements are observed after using auto-context steps. First, the label of the brain voxels considered as non-brain by the first networks in the middle of the brain voxels (i.e. false negatives) were changed to brain voxels (yellow arrows). Second, the very small number of the non-brain voxels considered as brain voxels in the first networks (white arrows) were changed to non-brain voxels. Third, the auto-context algorithm slightly pushed the edges of the brain to the outside (cyan arrows). These three improvements resulted in remarkable improvement in sensitivity at the cost of only a slight decrease in specificity in this case. The result is a significant boost in segmentation accuracy also shown by a significant increase in the Dice overlap coefficient.

It is worth noting that based on the data in Tables~\ref{table:Public} and~\ref{table:Fetal} the FCN (Auto-U-net) performed slightly better than the voxelwise CNN (Auto-2.5D-CNN) for the LPBA40 and OASIS datasets, but the voxelwise CNN outperformed FCN for the fetal MRI data. Our explanation is that there was higher level of commonality in shape and features of the samples in the LPBA40 and OASIS benchmark datasets compared to the fetal MRI dataset. This information was learned by the FCN, resulting in better performance compared to the voxelwise approach. For the fetal brain images that were arbitrarily located and oriented in the image space and surrounded by various structures, global geometric features were less important, and the voxelwise network performed better than the FCN as it learned and relied on 3D local image features.

\begin{figure*}
\centering
\includegraphics[width=\textwidth]{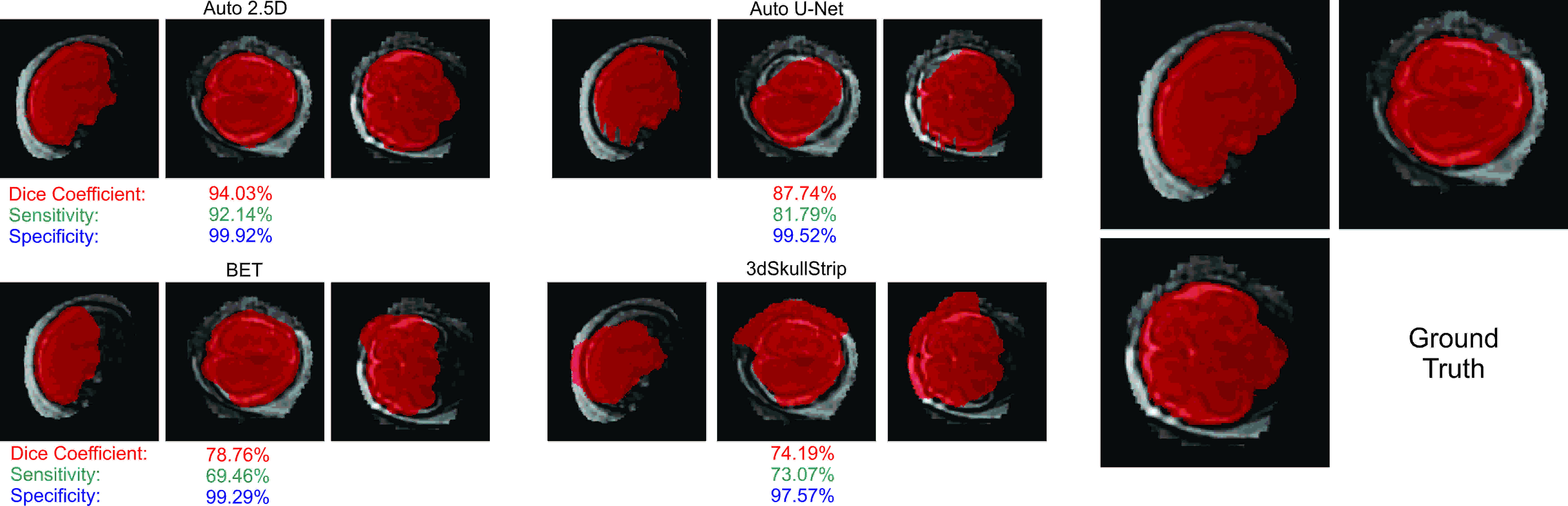}
\caption{Predicted masks overlaid on the reconstructed fetal brain MRI for a challenging case with decent image reconstruction quality and intensity non-uniformity due to B1 field inhomogeneity; the top images show the predicted brain masks by Auto-Net using 2.5D-CNN (left) and U-net (right). The bottom left and right images show the predicted brain masks using BET and 3dSkullStrip, respectively. The right image shows the ground truth manual segmentation. As can be seen, fetal brains can be in non-standard arbitrary orientations. Moreover, the fetal head may be surrounded by different tissue or organs. Despite all these challenges, the Auto-2.5D CNN performed well and much better than the other methods in this case. The Dice coefficient, sensitivity, and specificity, calculated based on the ground truth, are shown underneath each image in this figure.}
\label{fig:Challenge}
\end{figure*}

Figure~\ref{fig:Challenge} shows an example of a challenging fetal MRI case, where the voxelwise approach (Auto 2.5D) performed much better than the FCN approach (Auto U-net) as well as the other methods (BET and 3dSkullStrip). As can be seen from both Figures~\ref{fig:FetalSample} and~\ref{fig:Challenge}, fetal brains can be in non-standard arbitrary orientations, and the fetal head may be surrounded by different tissue or organs such as the amniotic fluid, uterus wall or placenta, or other fetal body parts such as hands or feet, or the umbilical cord. Despite the challenges raised, our Auto-Net methods, in particular the voxelwise CNN performed significantly better than the other methods in this application.

\begin{figure*}
    \centering
    \includegraphics[width=\textwidth]{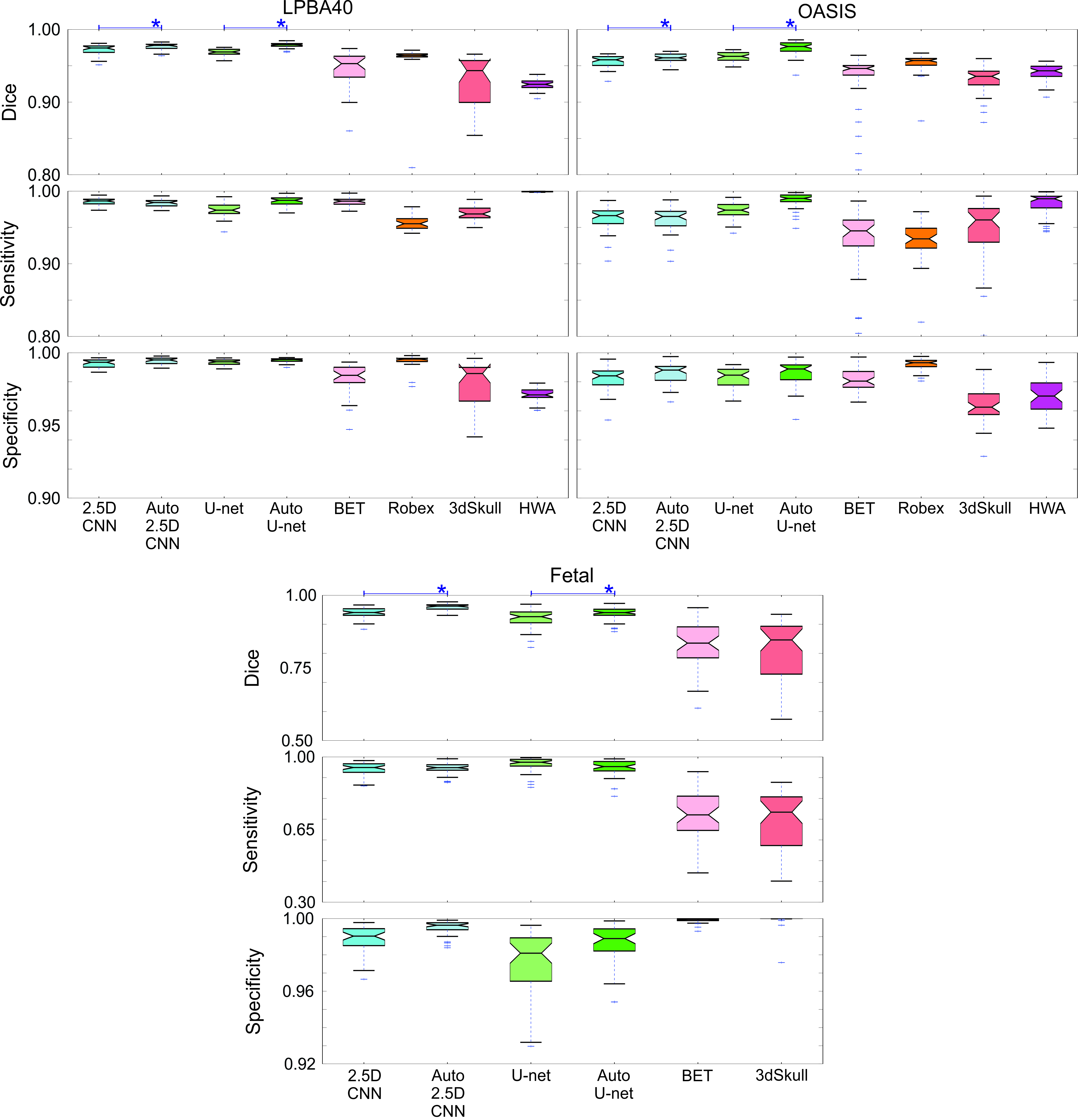}
    \caption{Evaluation scores (Dice, sensitivity, and specificity) for three data sets (LPBA40, OASIS, and fetal MRI). Median is displayed in boxplots; blue crosses represent outliers outside 1.5 times the interquartile range of the upper and lower quartiles, respectively. For the fetal dataset the registration-based algorithms were removed due to their poor performance. Those algorithms were not meant to work for images of this kind with non-standard geometry. Overall, these results show that our methods (Auto-Nets: Auto 2.5D and Auto U-net) made a very good trade-off between sensitivity and specificity and generated the highest Dice coefficients among all methods including the PCNN \cite{kleesiek2016deep}. The performance of Auto-Nets was consistently superior in the fetal MRI application where the other methods performed poorly due to the non-standard image geometry and features. Using Auto-context algorithm showed significant increase in Dice coefficients in both voxelwise and FCN style networks.}
    \label{fig:plotBox}
\end{figure*}

Figure~\ref{fig:plotBox} shows the box plots of the Dice coefficient, sensitivity, and specificity of the different algorithms on all three datasets. Among the non-CNN methods Robex performed well and was comparable to the 2.5D-CNN on the benchmark datasets, but could not be used reliably in the fetal dataset because of the geometric assumptions and the use of an atlas. On the other hand, BET and 3dSkullStrip had more relaxed assumptions thus could be used, albeit with limited accuracy. It should be noted that none of these methods were designed and tested for fetal brain MRI, so it was not expected that they worked well under the conditions of this dataset. In all datasets, Auto-Nets performed significantly better than all other methods as the auto-context significantly improved the results of both CNN architectures (2.5D and U-net).

Paired t-test was used to compare the results of different algorithms. The Dice coefficient of the proposed algorithm, Auto-Net (both Auto-2.5D and Auto-U-net), was significantly higher than BET, 3dSkullStrip, Robex, and HWA for LPBA40 and OASIS datasets at $\alpha$ threshold of 0.001 ($p<0.001$). Moreover, it revealed significant differences ($p<0.001$) between the Dice coefficient of the proposed algorithm (Auto-2.5D and Auto-U-net) with BET and 3dSkullStrip in fetal MRI. Paired t-test also showed significant improvement in the Dice coefficients obtained from the voxelwise network and the FCN through the use of the auto-context algorithm (i.e. Auto-2.5D vs. 2.5D and Auto-U-net vs. U-net).

\begin{figure}
    \centering
    \includegraphics[width=0.9\columnwidth]{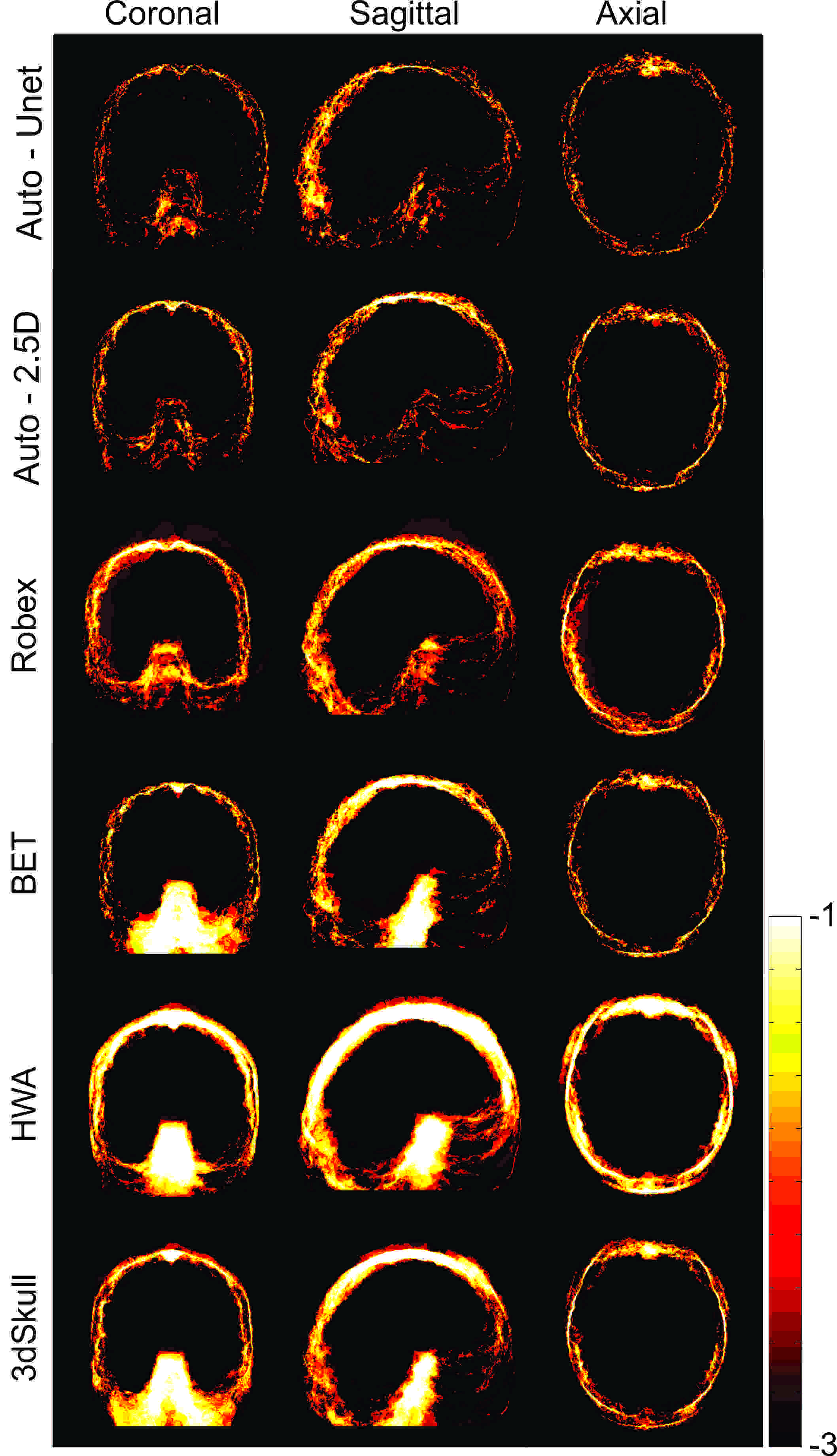}
    \caption{Logarithmic-scale absolute error maps of brain extraction obtained from six algorithms on the LPBA40 dataset. This analysis shows that Auto-Nets performed much better than the other methods in this dataset.}
    \label{fig:Error}
\end{figure}

Figure~\ref{fig:Error} shows logarithmic-scale average absolute error heat maps of the different algorithms on the LPBA40 dataset in the MNI atlas space~\cite{fonov2011unbiased}. These maps show where most errors occurred for each algorithm, and indicate that the Auto-Nets performed much better than the other methods in this dataset.

Table~\ref{table:Time} shows the average testing time (in seconds) for each dataset and each algorithm. It should be mentioned that the testing time for all the CNN-based methods including the PCNN were measured on GPUs, whereas the testing time for all non-CNN based methods were measured on multi-core CPUs, therefore this data does not directly compare the computational cost of different algorithms. It is also noteworthy that by using fully convolutional layers instead of fully connected layers in the 2.5D CNN architecture the testing time was decreased by a factor of almost 15 fold. Nonetheless, the FCN U-net is still significantly faster.

\begin{table}[h!]
\small
\centering
 \begin{tabular}{|c||c|c|c|} 
 \hline
Method & LPBA40 & OASIS & Fetal \\
 \hline
Auto-U-net & 10.03  & 22.85 & 14.11 \\ 
 \hline
U-net & 4.57 & 11.36 & 6.87 \\
 \hline
Auto-2.5D-CNN & 794.42 & 641.26 & 501.73 \\ 
 \hline
2.5D-CNN & 396.23 & 320.12 & 244.9 \\ 
 \hline
PCNN & 36.51 & 40.99 & - \\ 
 \hline
BET & \textbf{2.04} & \textbf{1.96} & \textbf{1.62} \\
\hline
3dSkullStrip & 130.4 & 119.12 & 82.72 \\
\hline
Robex & 52.10 & 63.25 & - \\
\hline
HWA & 18.73 & 13.42 & -\\ 
\hline
\end{tabular}
\caption{Average runtimes (seconds) of the methods compared in this study: the non-CNN methods were tested on an Intel(R) Core(TM) i7-5930K CPU with 3.50 GHz and 64 GB RAM for all data sets (LPBA40, OASIS and Fetal). The CNN-based methods were tested on an NVIDIA GeForce GTX 1080 (Pascal architecture). The PCNN timings are based on those reported in \cite{kleesiek2016deep} using an NVIDIA Titan GPU with Kepler architecture.}
\label{table:Time}

\end{table}

\section{Discussion}

Our proposed auto-context convolutional neural networks outperformed the recent deep learning method~\cite{kleesiek2016deep} and four widely-used brain extraction techniques that were continuously evolved and improved over the past decade due to the significant demand for accurate and reliable automated brain extraction in the neuroscience and neuroimaging communities.

We achieved the highest Dice coefficients as well as a good sensitivity-specificity trade-off among the techniques examined in this paper. This was achieved by using the auto-context algorithm and FCN approach together for standard datasets and auto-context with multiple patch sizes as well as context information in a voxelwise CNN architecture.

While the auto-context FCN based on U-net was much faster than the auto-context voxelwise network, it performed only slightly better for the benchmark datasets. On the other hand, the auto-context voxelwise network performed much better than the auto-context FCN in the very challenging fetal MRI brain extraction problem. The auto-context algorithm dramatically improved the performance of both networks.

We trained and examined efficient voxelwise and FCN Auto-Nets in this paper. Extensions to 3D networks is analytically straightforward; but the 3D counterparts are typically more demanding on computational resources, in particular memory. Generally, in voxelwise networks each voxel is considered as an independent sample to be classified. A window or different-sized windows around voxels are chosen as features and the network is trained using those features. Kleesiek et al.~\cite{kleesiek2016deep} used one cube with constant window size around each voxel. Moeskops et al.~\cite{moeskops2016automatic} used different window sizes around voxels but in 2D views. The main reason that previous studies did not use both approaches together, is that the number of parameters increases significantly with 3D convolutional kernels, especially when different, typically large window sizes are used. Such a network can easily consume more memory than what is available on most workstation GPUs. Our 2.5D network made a good trade-off in this regards.

To compare our 2.5D network (which consists of three 2D pathway networks) with its 3D counterpart, we calculate the number of parameters: The 2.5D network contains 68.81 million parameters whereas its 3D counterpart contains 793.84 million parameters. With direct implementation with a small batch size of 1, the 3D counterpart of our CNN consumes more than 40GB of GPU memory. On the other hand, in our 2.5D network architecture we efficiently used a batch size of 64. The nearest 3D counterpart of our network with similar memory usage contained two cubes with window sizes of 15 and 41. We tested this network on the LPBA40 dataset and observed $1.5\%$ decrease in average Dice coefficients while the average testing time increased by a factor of 1.5. This architecture contained 154 million parameters. We also systematically evaluated the effect of the three pathways and different window sizes. To this end, we trained and tested networks with only one pathway in each plane. While the testing times were decreased by a factor of 4, we observed significant decrease in average Dice coefficients, at $2.8-4.3\%$. We also observed significant decrease in average Dice coefficients by using single window sizes instead of using different window sizes (i.e. $5\%, 2.1\%$, and $0.9\%$ drop in the Dice coefficients for window sizes of 15, 25, and 51, respectively).

With Auto-Net we overcome one of the persisting challenges in fetal brain MRI processing. The extraction of fetal brain from reconstructed fetal MRI previously required a significant amount of work to correct the masks provided by BET or other level set whole brain segmentation techniques~\cite{gholipour2012multi,gholipour2011fetal}. Atlas-based segmentation methods heavily rely on image registration which involves time-consuming search and optimization to match the arbitrary orientation of images~\cite{taimouri2015template}, followed by deformable registration to age-matched templates~\cite{tourbier2017automated}, or patch-based label propagation~\cite{wright2014automatic}, which are also time consuming and difficult due to the presence of residual non-brain tissue after initial alignments. Most of the work in the literature focused on brain detection and localization in original fetal brain MRI scans, mainly to improve automated motion correction and reconstruction, e.g.~\cite{kainz2014fast,keraudren2014automated,keraudren2015automated}. While accurate bounding boxes are detected around the fetal brain by these methods, leading to improved motion correction~\cite{keraudren2014automated}, the estimated brain masks are not exact and consequently the reconstructed images involve significant non-brain tissue. Therefore accurate brain extraction is critically needed after reconstruction. Rather than being dependent on difficult and time-consuming image registration processes, the Auto-Net fetal brain extractions, proposed here, work at the voxel level to mask the fetal brains and prepare them for registration to an atlas space~\cite{gholipour2017normative} for further analysis. Brain masks are also useful in other processing tasks, such as intensity non-uniformity correction~\cite{tustison2010n4itk}, which poses significant challenges in fetal MRI as can be seen in Figure~\ref{fig:Challenge}.

In comparison with other methods, the features in CNN-based methods are learnt through the training step and no hand-crafted features are needed. After training, these methods are fast in testing. We noted that these methods do not use image registration nor do they make assumptions about global image geometry. Rather, the networks learn to classify voxels based on local and shape image features. An inherent assumption in such learning-based methods is that a suitable training set is available. This is a strict assumption both in terms of the existence of the training set and in that any new test image should have the same feature distribution as the training set. We used one modality in this study. It is expected that if multiple modalities, such as T1-weighted, T2-weighted, FLAIR,  or CT images along with their training data are available and used, they result in increased accuracy. The only change in the architecture will be the additional third dimension of the kernel of the first convolutional layer.

\section{Conclusion}

We developed and evaluated auto-context convolutional neural networks with two different architectures (a voxelwise network with three parallel 2D pathways, and a FCN style U-net) for whole-brain segmentation in 3D MRI. The auto-context CNNs outperformed a recent deep learning method and four widely-used brain extraction methods in two publicly available benchmark datasets and in the very challenging problem of extracting fetal brain from reconstructed fetal MRI. Unlike the current highly evolved brain extraction methods that use a combination of surface models, surface evolutions, and edge and intensity features, CNN-based methods do not use image registration or assume global geometric features such as certain orientations, but require suitable training data.

\appendices
\section{}
\label{appendix}

\textbf{Theorem 1. } The cross-entropy cost function in Algorithm~\ref{algorithm} monotonically decreases during the training.

\begin{proof}
To show that the cross-entropy cost function decreases monotonically, we show that the cost at each level will be smaller or at least equal to the cost at previous level. At the arbitrary step $t$, 
\begin{equation}
\begin{split}
    H_t & = - \sum_i log \ p_{(t),i}(y_i) \\ 
    & = - \sum_i log \ p_{(t)}(y_i|(X^{(j)}(N_i),p_{(t-1)}(N_i))) 
\end{split}
\end{equation}
and 
\begin{equation}
\begin{split}
    H_{t-1} & = - \sum_i log \ p_{(t-1),i}(y_i) 
\end{split}
\end{equation}
Also, note that the posterior probability is:
\begin{equation}
    p_{(t)}(y_i=k|X(N(i),p_{(t-1)}(N_i)) = \frac{e^{f_{y_k}(X(N_i),p_{(t-1)}(N_i))}}{\sum_ce^{f_{y_c}(X(N_i),p_{(t-1)}(N_i))}}
    \label{Autosoftmax}
\end{equation} 
Using $f_{y_k}(X(N_i),p_{(t-1)}(N_i)) = log \ p_{(t-1),i}(y_i)$ cross-entropy in level $t$ will be equal to cross-entropy in level $t-1$. Since, during the training in step $t$ we are minimizing the cross entropy cost function, ${p}_{(t),i}(y_i)$ should at least work better than ${p}_{(t-1),i}(y_i)$. Therefore:

\begin{equation}
    H_{(t)} \leqslant H_{(t-1)}
\end{equation}

\end{proof}

\section*{Acknowledgment}
This study was supported in part by the National Institute of Biomedical Imaging and Bioengineering of the National Institutes of Health (NIH) grant R01 EB018988. The content of this work is solely the responsibility of the authors and does not necessarily represent the official views of the NIH.

\ifCLASSOPTIONcaptionsoff
  \newpage
\fi



\bibliographystyle{IEEEtran}
%
\bibliography{Ref}

\begin{thebibliography}{10}
\providecommand{\url}[1]{#1}
\csname url@samestyle\endcsname
\providecommand{\newblock}{\relax}
\providecommand{\bibinfo}[2]{#2}
\providecommand{\BIBentrySTDinterwordspacing}{\spaceskip=0pt\relax}
\providecommand{\BIBentryALTinterwordstretchfactor}{4}
\providecommand{\BIBentryALTinterwordspacing}{\spaceskip=\fontdimen2\font plus
\BIBentryALTinterwordstretchfactor\fontdimen3\font minus
  \fontdimen4\font\relax}
\providecommand{\BIBforeignlanguage}[2]{{%
\expandafter\ifx\csname l@#1\endcsname\relax
\typeout{** WARNING: IEEEtran.bst: No hyphenation pattern has been}%
\typeout{** loaded for the language `#1'. Using the pattern for}%
\typeout{** the default language instead.}%
\else
\language=\csname l@#1\endcsname
\fi
#2}}
\providecommand{\BIBdecl}{\relax}
\BIBdecl

\bibitem{makropoulos2014automatic}
A.~Makropoulos, I.~S. Gousias, C.~Ledig, P.~Aljabar, A.~Serag, J.~V. Hajnal,
  A.~D. Edwards, S.~J. Counsell, and D.~Rueckert, ``Automatic whole brain {MRI}
  segmentation of the developing neonatal brain,'' \emph{IEEE transactions on
  medical imaging}, vol.~33, no.~9, pp. 1818--1831, 2014.

\bibitem{li2014mapping}
G.~Li, L.~Wang, F.~Shi, A.~E. Lyall, W.~Lin, J.~H. Gilmore, and D.~Shen,
  ``Mapping longitudinal development of local cortical gyrification in infants
  from birth to 2 years of age,'' \emph{The Journal of Neuroscience}, vol.~34,
  no.~12, pp. 4228--4238, 2014.

\bibitem{macdonald2000automated}
D.~MacDonald, N.~Kabani, D.~Avis, and A.~C. Evans, ``Automated {3D} extraction
  of inner and outer surfaces of cerebral cortex from {MRI},''
  \emph{NeuroImage}, vol.~12, no.~3, pp. 340--356, 2000.

\bibitem{clouchoux2012quantitative}
C.~Clouchoux, D.~Kudelski, A.~Gholipour, S.~K. Warfield, S.~Viseur,
  M.~Bouyssi-Kobar, J.-L. Mari, A.~C. Evans, A.~J. Du~Plessis, and
  C.~Limperopoulos, ``Quantitative in vivo {MRI} measurement of cortical
  development in the fetus,'' \emph{Brain Structure and Function}, vol. 217,
  no.~1, pp. 127--139, 2012.

\bibitem{de2015deep}
A.~de~Brebisson and G.~Montana, ``Deep neural networks for anatomical brain
  segmentation,'' in \emph{Proceedings of the IEEE Conference on Computer
  Vision and Pattern Recognition Workshops}, 2015, pp. 20--28.

\bibitem{wachinger2017deepnat}
C.~Wachinger, M.~Reuter, and T.~Klein, ``{DeepNAT}: Deep convolutional neural
  network for segmenting neuroanatomy,'' \emph{NeuroImage}, 2017.

\bibitem{pereira2016brain}
S.~Pereira, A.~Pinto, V.~Alves, and C.~A. Silva, ``Brain tumor segmentation
  using convolutional neural networks in {MRI} images,'' \emph{IEEE
  transactions on medical imaging}, vol.~35, no.~5, pp. 1240--1251, 2016.

\bibitem{havaei2016brain}
M.~Havaei, A.~Davy, D.~Warde-Farley, A.~Biard, A.~Courville, Y.~Bengio, C.~Pal,
  P.-M. Jodoin, and H.~Larochelle, ``Brain tumor segmentation with deep neural
  networks,'' \emph{Medical Image Analysis}, 2016.

\bibitem{brosch2016deep}
T.~Brosch, L.~Y. Tang, Y.~Yoo, D.~K. Li, A.~Traboulsee, and R.~Tam, ``Deep {3D}
  convolutional encoder networks with shortcuts for multiscale feature
  integration applied to multiple sclerosis lesion segmentation,'' \emph{IEEE
  transactions on medical imaging}, vol.~35, no.~5, pp. 1229--1239, 2016.

\bibitem{kamnitsas2016efficient}
K.~Kamnitsas, C.~Ledig, V.~F. Newcombe, J.~P. Simpson, A.~D. Kane, D.~K. Menon,
  D.~Rueckert, and B.~Glocker, ``Efficient multi-scale {3D} {CNN} with fully
  connected {CRF} for accurate brain lesion segmentation,'' \emph{arXiv
  preprint arXiv:1603.05959}, 2016.

\bibitem{zhang2015deep}
W.~Zhang, R.~Li, H.~Deng, L.~Wang, W.~Lin, S.~Ji, and D.~Shen, ``Deep
  convolutional neural networks for multi-modality isointense infant brain
  image segmentation,'' \emph{NeuroImage}, vol. 108, pp. 214--224, 2015.

\bibitem{kleesiek2016deep}
J.~Kleesiek, G.~Urban, A.~Hubert, D.~Schwarz, K.~Maier-Hein, M.~Bendszus, and
  A.~Biller, ``Deep {MRI} brain extraction: a {3D} convolutional neural network
  for skull stripping,'' \emph{NeuroImage}, vol. 129, pp. 460--469, 2016.

\bibitem{smith2002fast}
S.~M. Smith, ``Fast robust automated brain extraction,'' \emph{Human brain
  mapping}, vol.~17, no.~3, pp. 143--155, 2002.

\bibitem{jenkinson2005bet2}
M.~Jenkinson, M.~Pechaud, and S.~Smith, ``{BET2}: {MR}-based estimation of
  brain, skull and scalp surfaces,'' in \emph{11th annual meeting of the
  organization for human brain mapping}, vol.~17, 2005, p. 167.

\bibitem{cox1996afni}
R.~W. Cox, ``{AFNI}: software for analysis and visualization of functional
  magnetic resonance neuroimages,'' \emph{Computers and Biomedical research},
  vol.~29, no.~3, pp. 162--173, 1996.

\bibitem{lin2003hybrid}
G.~Lin, U.~Adiga, K.~Olson, J.~F. Guzowski, C.~A. Barnes, and B.~Roysam, ``A
  hybrid {3D} watershed algorithm incorporating gradient cues and object models
  for automatic segmentation of nuclei in confocal image stacks,''
  \emph{Cytometry Part A}, vol.~56, no.~1, pp. 23--36, 2003.

\bibitem{iglesias2011robust}
J.~E. Iglesias, C.-Y. Liu, P.~M. Thompson, and Z.~Tu, ``Robust brain extraction
  across datasets and comparison with publicly available methods,'' \emph{IEEE
  transactions on medical imaging}, vol.~30, no.~9, pp. 1617--1634, 2011.

\bibitem{moeskops2016automatic}
P.~Moeskops, M.~A. Viergever, A.~M. Mendrik, L.~S. de~Vries, M.~J. Benders, and
  I.~I{\v{s}}gum, ``Automatic segmentation of {MR} brain images with a
  convolutional neural network,'' \emph{IEEE transactions on medical imaging},
  vol.~35, no.~5, pp. 1252--1261, 2016.

\bibitem{ronneberger2015u}
O.~Ronneberger, P.~Fischer, and T.~Brox, ``U-net: Convolutional networks for
  biomedical image segmentation,'' in \emph{International Conference on Medical
  Image Computing and Computer-Assisted Intervention}.\hskip 1em plus 0.5em
  minus 0.4em\relax Springer, 2015, pp. 234--241.

\bibitem{lafferty2001conditional}
J.~Lafferty, A.~McCallum, F.~Pereira \emph{et~al.}, ``Conditional random
  fields: Probabilistic models for segmenting and labeling sequence data,'' in
  \emph{Proceedings of the eighteenth international conference on machine
  learning, ICML}, vol.~1, 2001, pp. 282--289.

\bibitem{chen2014semantic}
L.-C. Chen, G.~Papandreou, I.~Kokkinos, K.~Murphy, and A.~L. Yuille, ``Semantic
  image segmentation with deep convolutional nets and fully connected {CRFs},''
  \emph{arXiv preprint arXiv:1412.7062}, 2014.

\bibitem{valverde2017improving}
S.~Valverde, M.~Cabezas, E.~Roura, S.~Gonz{\'a}lez-Vill{\`a}, D.~Pareto, J.~C.
  Vilanova, L.~Rami{\'o}-Torrent{\`a}, {\`A}.~Rovira, A.~Oliver, and
  X.~Llad{\'o}, ``Improving automated multiple sclerosis lesion segmentation
  with a cascaded {3D} convolutional neural network approach,''
  \emph{NeuroImage}, 2017.

\bibitem{tu2010auto}
Z.~Tu and X.~Bai, ``Auto-context and its application to high-level vision tasks
  and {3D} brain image segmentation,'' \emph{IEEE Transactions on Pattern
  Analysis and Machine Intelligence}, vol.~32, no.~10, pp. 1744--1757, 2010.

\bibitem{ciresan2012deep}
D.~Ciresan, A.~Giusti, L.~M. Gambardella, and J.~Schmidhuber, ``Deep neural
  networks segment neuronal membranes in electron microscopy images,'' in
  \emph{Advances in neural information processing systems}, 2012, pp.
  2843--2851.

\bibitem{long2015fully}
J.~Long, E.~Shelhamer, and T.~Darrell, ``Fully convolutional networks for
  semantic segmentation,'' in \emph{Proceedings of the IEEE Conference on
  Computer Vision and Pattern Recognition}, 2015, pp. 3431--3440.

\bibitem{shelhamer2017fully}
E.~Shelhamer, J.~Long, and T.~Darrell, ``Fully convolutional networks for
  semantic segmentation,'' \emph{IEEE transactions on pattern analysis and
  machine intelligence}, vol.~39, no.~4, pp. 640--651, 2017.

\bibitem{sermanet2013overfeat}
P.~Sermanet, D.~Eigen, X.~Zhang, M.~Mathieu, R.~Fergus, and Y.~LeCun,
  ``Overfeat: Integrated recognition, localization and detection using
  convolutional networks,'' \emph{arXiv preprint arXiv:1312.6229}, 2013.

\bibitem{kingma2014adam}
D.~Kingma and J.~Ba, ``Adam: A method for stochastic optimization,''
  \emph{arXiv preprint arXiv:1412.6980}, 2014.

\bibitem{shattuck2009online}
D.~W. Shattuck, G.~Prasad, M.~Mirza, K.~L. Narr, and A.~W. Toga, ``Online
  resource for validation of brain segmentation methods,'' \emph{NeuroImage},
  vol.~45, no.~2, pp. 431--439, 2009.

\bibitem{marcus2007open}
D.~S. Marcus, T.~H. Wang, J.~Parker, J.~G. Csernansky, J.~C. Morris, and R.~L.
  Buckner, ``Open access series of imaging studies {(OASIS)}: cross-sectional
  mri data in young, middle aged, nondemented, and demented older adults,''
  \emph{Journal of cognitive neuroscience}, vol.~19, no.~9, pp. 1498--1507,
  2007.

\bibitem{yushkevich2006user}
P.~A. Yushkevich, J.~Piven, H.~C. Hazlett, R.~G. Smith, S.~Ho, J.~C. Gee, and
  G.~Gerig, ``User-guided {3D} active contour segmentation of anatomical
  structures: significantly improved efficiency and reliability,''
  \emph{Neuroimage}, vol.~31, no.~3, pp. 1116--1128, 2006.

\bibitem{gholipour2010robust}
A.~Gholipour, J.~A. Estroff, and S.~K. Warfield, ``Robust super-resolution
  volume reconstruction from slice acquisitions: application to fetal brain
  {MRI},'' \emph{IEEE transactions on medical imaging}, vol.~29, no.~10, pp.
  1739--1758, 2010.

\bibitem{kainz2015fast}
B.~Kainz, M.~Steinberger, W.~Wein, M.~Kuklisova-Murgasova, C.~Malamateniou,
  K.~Keraudren, T.~Torsney-Weir, M.~Rutherford, P.~Aljabar, J.~V. Hajnal
  \emph{et~al.}, ``Fast volume reconstruction from motion corrupted stacks of
  {2D} slices,'' \emph{IEEE transactions on medical imaging}, vol.~34, no.~9,
  pp. 1901--1913, 2015.

\bibitem{fonov2011unbiased}
V.~Fonov, A.~C. Evans, K.~Botteron, C.~R. Almli, R.~C. McKinstry, D.~L.
  Collins, B.~D.~C. Group \emph{et~al.}, ``Unbiased average age-appropriate
  atlases for pediatric studies,'' \emph{NeuroImage}, vol.~54, no.~1, pp.
  313--327, 2011.

\bibitem{gholipour2012multi}
A.~Gholipour, A.~Akhondi-Asl, J.~A. Estroff, and S.~K. Warfield, ``Multi-atlas
  multi-shape segmentation of fetal brain {MRI} for volumetric and morphometric
  analysis of ventriculomegaly,'' \emph{Neuroimage}, vol.~60, no.~3, pp.
  1819--1831, 2012.

\bibitem{gholipour2011fetal}
A.~Gholipour, J.~A. Estroff, C.~E. Barnewolt, S.~A. Connolly, and S.~K.
  Warfield, ``Fetal brain volumetry through {MRI} volumetric reconstruction and
  segmentation,'' \emph{International journal of computer assisted radiology
  and surgery}, vol.~6, no.~3, pp. 329--339, 2011.

\bibitem{taimouri2015template}
V.~Taimouri, A.~Gholipour, C.~Velasco-Annis, J.~A. Estroff, and S.~K. Warfield,
  ``A template-to-slice block matching approach for automatic localization of
  brain in fetal {MRI},'' in \emph{Biomedical Imaging (ISBI), 2015 IEEE 12th
  International Symposium on}.\hskip 1em plus 0.5em minus 0.4em\relax IEEE,
  2015, pp. 144--147.

\bibitem{tourbier2017automated}
S.~Tourbier, C.~Velasco-Annis, V.~Taimouri, P.~Hagmann, R.~Meuli, S.~K.
  Warfield, M.~B. Cuadra, and A.~Gholipour, ``Automated template-based brain
  localization and extraction for fetal brain {MRI} reconstruction,''
  \emph{NeuroImage}, 2017.

\bibitem{wright2014automatic}
R.~Wright, V.~Kyriakopoulou, C.~Ledig, M.~A. Rutherford, J.~V. Hajnal,
  D.~Rueckert, and P.~Aljabar, ``Automatic quantification of normal cortical
  folding patterns from fetal brain {MRI},'' \emph{NeuroImage}, vol.~91, pp.
  21--32, 2014.

\bibitem{kainz2014fast}
B.~Kainz, K.~Keraudren, V.~Kyriakopoulou, M.~Rutherford, J.~V. Hajnal, and
  D.~Rueckert, ``Fast fully automatic brain detection in fetal {MRI} using
  dense rotation invariant image descriptors,'' in \emph{Biomedical Imaging
  (ISBI), 2014 IEEE 11th International Symposium on}.\hskip 1em plus 0.5em
  minus 0.4em\relax IEEE, 2014, pp. 1230--1233.

\bibitem{keraudren2014automated}
K.~Keraudren, M.~Kuklisova-Murgasova, V.~Kyriakopoulou, C.~Malamateniou, M.~A.
  Rutherford, B.~Kainz, J.~V. Hajnal, and D.~Rueckert, ``Automated fetal brain
  segmentation from {2D MRI} slices for motion correction,'' \emph{NeuroImage},
  vol. 101, pp. 633--643, 2014.

\bibitem{keraudren2015automated}
K.~Keraudren, B.~Kainz, O.~Oktay, V.~Kyriakopoulou, M.~Rutherford, J.~V.
  Hajnal, and D.~Rueckert, ``Automated localization of fetal organs in {MRI}
  using random forests with steerable features,'' in \emph{International
  Conference on Medical Image Computing and Computer-Assisted
  Intervention}.\hskip 1em plus 0.5em minus 0.4em\relax Springer, 2015, pp.
  620--627.

\bibitem{gholipour2017normative}
A.~Gholipour, C.~K. Rollins, C.~Velasco-Annis, A.~Ouaalam, A.~Akhondi-Asl,
  O.~Afacan, C.~M. Ortinau, S.~Clancy, C.~Limperopoulos, E.~Yang, J.~Estroff,
  and S.~K. Warfield, ``A normative spatiotemporal {MRI} atlas of the fetal
  brain for automatic segmentation and analysis of early brain growth,''
  \emph{Scientific Reports}, vol.~7, no.~1, p. 476, 2017.

\bibitem{tustison2010n4itk}
N.~J. Tustison, B.~B. Avants, P.~A. Cook, Y.~Zheng, A.~Egan, P.~A. Yushkevich,
  and J.~C. Gee, ``{N4ITK: improved N3 bias correction},'' \emph{IEEE
  transactions on medical imaging}, vol.~29, no.~6, pp. 1310--1320, 2010.

\end{thebibliography}

%








\end{document}